# SENSING URBAN LAND-USE PATTERNS BY INTEGRATING GOOGLE TENSORFLOW AND SCENE-CLASSIFICATION MODELS

Yao Yao [a], Haolin Liang [a], Xia Li [b, *], Jinbao Zhang [a], Jialv He [a]

[a] Guangdong Key Laboratory for Urbanization and Geo-simulation, Sun Yat-sen University, School of Geography and Planning, Guangzhou, Guangdong province, China – (whuyao, borisliang, kampau, hesysugis)@foxmail.com

[b] Guangdong Key Laboratory for Urbanization and Geo-simulation, Sun Yat-sen University, School of Geography and Planning, Guangzhou, Guangdong province, China – lixia@mail.sysu.edu.cn

**ABSTRACT:**

With the rapid progress of China's urbanization, research on the automatic detection of land-use patterns in Chinese cities is of substantial importance. Deep learning is an effective method to extract image features. To take advantage of the deep-learning method in detecting urban land-use patterns, we applied a transfer-learning-based remote-sensing image approach to extract and classify features. Using the Google Tensorflow framework, a powerful convolution neural network (CNN) library was created. First, the transferred model was previously trained on ImageNet, one of the largest object-image data sets, to fully develop the model's ability to generate feature vectors of standard remote-sensing land-cover data sets (UC Merced and WHU-SIRI). Then, a random-forest-based classifier was constructed and trained on these generated vectors to classify the actual urban land-use pattern on the scale of traffic analysis zones (TAZs). To avoid the multi-scale effect of remote-sensing imagery, a large random patch (LRP) method was used. The proposed method could efficiently obtain acceptable accuracy (OA = 0.794, Kappa = 0.737) for the study area. In addition, the results show that the proposed method can effectively overcome the multi-scale effect that occurs in urban land-use classification at the irregular land-parcel level. The proposed method can help planners monitor dynamic urban land use and evaluate the impact of urban-planning schemes.

## 1. INTRODUCTION

Land-use and land-cover (LULC) information plays an essential role in many fields, such as environmental monitoring, urban planning and governmental management (Liu et al., 2017; Lu and Weng, 2006; Williamson et al., 2010; Zhang et al., 2015). The transformation of urban functional structures and urban land-use patterns is affected not only by top-down influences from government but also by bottom-up influences from residents (Chen et al., 2017). With economic development, the transformation of China's urban land-use pattern is accelerating. Traditional urban land-use surveying and mapping methods are costly in time and money. Thus, it is important to develop a quick and accurate method to sense urban land-use patterns via remote-sensing imagery.

In recent years, mainstream studies on recognizing urban land-use patterns have been based on high spatial resolution (HSR) image classification (Bratasanu et al., 2011; Chen et al., 2013; Wen et al., 2016; Zhong et al., 2015). HSR images contain a substantial amount of natural-physical geospatial information, which is widely used in the object-oriented classification (OOC) method to extract urban land-use information (Bratasanu et al., 2011; Durand et al., 2007). The traditional OOC method classifies urban land-use or land-cover types by distinguishing multi-dimensional features, including spectrum, texture and shape, of various land parcels.

Urban land-use patterns are strongly correlated with government policies and resident activities (Hu et al., 2016; Liu et al., 2017; Pei et al., 2014). The inner structure of urban land use is complicated wherever the so-called 'semantic gap' requires bridging (Bratasanu et al., 2011; Liu et al., 2017). That is, recognizing urban land-use conditions requires consideration of the composition and structure of internal ground objects. Such consideration requires a substantial human effort to obtain high-level semantic information and then to classify the urban land-use type via scene classification (Liu et al., 2017; Zhong et al., 2015). The state of the art of classifying urban land use via scene classification is represented by Zhong et al. (2015), who used a probabilistic topic model (PTM) to fuse spectral and textural features from HSR images. However, most remote-sensing, imagery-based urban land-use detection methods are implemented on land parcels with regular shapes. However, to our knowledge, the fundamental unit of land use in most cities is the irregular land parcel or traffic analysis zones planned by governments. Irregular parcels cause uncertainty when the features of remotely sensed images are extracted (Liu et al., 2017; Long and Liu, 2015; Yao et al., 2016). Specifically, with an increase in image spatial resolution, the spatial structure of ground components exhibits a recognizable heterogeneity, which results in a "multi-scale effect" problem when such features are classified (Zhong et al., 2016). This problem is a significant difficulty faced by traditional OOC methods.

Only a small number of studies have focused on fusing remote-sensing image data with multi-source social media data to classify urban land use (Liu et al., 2017; Yao et al., 2016). For example, Hu et al. (2015) adopted Landsat remote-sensing images and POI data to recognize urban land-use conditions (Hu et al., 2015). Liu et al. (2017) used Worldview-2 images and several social media data sources with probabilistic topic models (PTMs) to detect the urban land-use condition in Guangzhou and obtained relatively good results (Liu et al., 2017). However, the accuracy of multi-source fusion methods has not been thoroughly demonstrated. In particular, the influence of bias in social-media data must be further

---

* Corresponding author

considered (Zheng, 2015).

Probing more deeply into HSR images to obtain high-level semantic information to simulate the urban spatial transformation is a popular recent research approach (Yao et al., 2016; Zhong et al., 2016). Zhong et al. (2016) manipulated a deep-learning method and created HSR image subsets to solve the multi-scale problem in remote-sensing image classification (Zhong et al. 2016). Jean used the transfer-learning method to transfer the knowledge of a pre-trained, object-based image classifier to extract income distribution (Jean et al., 2016). As these studies indicate, applying deep learning in the field of remote sensing remains in the preliminary stage, and the integration of the method into urban land-use classification is even less advanced. However, several studies have demonstrated that the deep-learning method is effective in solving the 'semantic gap' problem in semantic classification because it avoids establishing a complex, rule-based classifier (Nogueira et al., 2016; Zhong et al., 2016). Thus, deep learning represents a highly promising approach to the study of urban land-use conditions and deserves further investigation.

This article introduces an urban land-use classification method that integrates deep learning and semantic models to classify irregular land parcels only using HSR images. First, to classify land-cover images, the proposed method uses public land-cover data sets (UC-Merced and WHU-SIRI) to transfer information from an object image-based deep-learning model (Inception v5). Then, with a retrained deep-learning model, we extract the class types of random patches inside the land parcels with irregular shapes and assemble a visual bag of words. Finally, we use a simple semantic model named TF-IDF to establish the semantic features of each land parcel and a random forest (RFA)-based classifier to classify the urban land-use conditions.

## 2. STUDY AREA AND DATA

Known as the Haizhu district, the study area is located in the central area of the city of Guangzhou, Guangdong Province (Figures 1 and 2). It has a total area of 102 km2 and a permanent population of approximately 1,010,500. Guangzhou, which is the capital of Guangdong Province and boasts a history of approximately 2,000 years, is the political, economic and cultural center of southern China. The Haizhu district is the main, central district of Guangzhou and exhibits a complicated urban functional structure and various land-use patterns. The land-use data for the study area at the land-parcel level consist of traffic analysis zones (TAZs) (Figure 1). Based on high spatial resolution (HSR) image data with a spatial resolution of 0.5 meters downloaded from Tianditu.cn (Figure 2) and field sampling, we classify 593 TAZs in the study area into 7 different dominated land-use types: Public management-services land (M), Industrial land (I), Green land (G), Commercial land (C), Residential land (R), Park land (P) and Urban village (U) (Liu et al., 2017).

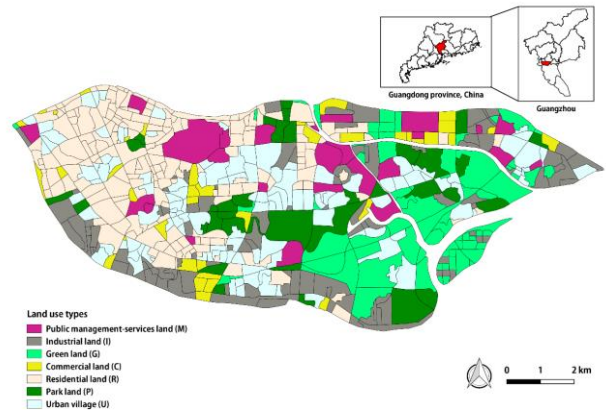

Figure 1. Urban land-use data obtained from manual interpretation at the land-parcel level for the case-study area: Haizhu district, Guangzhou, Guangdong Province.

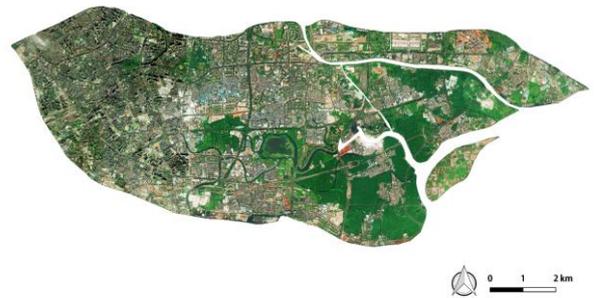

Figure 2. High spatial resolution (HSR) remote-sensing image of the study area provided by Worldview-2 satellite.

To use the high-accuracy object-based deep-learning model, we retrain the Google-introduced Inception v5 model to classify standard land-cover data sets: UC-Merced and WHU-SIRI. The UC-Merced land-cover data set (http://vision.ucmerced.edu/datasets/landuse.html) consists of 21 different types of land-cover remote-sensing image. Each class contains 100 different 2-meter-resolution remote-sensing images. Each image is 256*256 pixels, which means that the resolution of a single pixel is 1 foot. The 27 classes include agricultural land, airport, baseball court, beach, building, agricultural, airplane, baseball diamond, beach, buildings, chaparral, dense residential, forest, freeway, golf course, harbor, intersection, medium residential, mobile home park, overpass, parking lot, river, runway, sparse residential, storage tanks and tennis court. The UC-Merced image data are from the U.S. The WHU-SIRI land-use data set (http://www.lmars.whu.edu.cn/prof_web/zhongyanfei/e-code.html) consists of 12 different types of land-cover remote-sensing images. Each class contains 200 different 2-meter-resolution remote-sensing images, and each image is 200*200 pixels. These 12 classes include agriculture, commercial, harbor, idle land, industrial, meadow, overpass, park, pond, residential, river and water. The WHU-SIRI data were collected via Google Earth within the urban area of China.

## 3. METHOD

A flowchart of the proposed model is presented in Figure 3. This study aims to classify urban land use at the irregular land-parcel level by integrating deep-learning and scene-classification in the model. First, we transfer the information obtained using a pre-trained object image-based deep-learning model to classify land-cover images by retraining on standard land-cover data sets (UC-Merced and WHU-SIRI). Then, using the retrained model, we transform each multi-scale sample into a word according to land-cover type and count the word frequencies under the unit of land parcel (TAZ). Last, we use the TF-IDF algorithm to transform the word frequencies within each TAZ into semantic features and use the manually interpreted urban land-use types and semantic features to construct a random-forest (RFA)-based classifier. Subsequently, the model's accuracy is assessed.

$$\sigma(z)_j = \frac{e^{z_j}}{\sum_{k=1}^{K} e^{z_k}} \quad j = 1, \ldots, K \quad (1)$$

In Equation (1), $\sigma(z)_j$ represents the probability of samples being classified into class $j$, and $z_j$ represents each element from the high-dimensional vector produced by the Bottleneck layer.

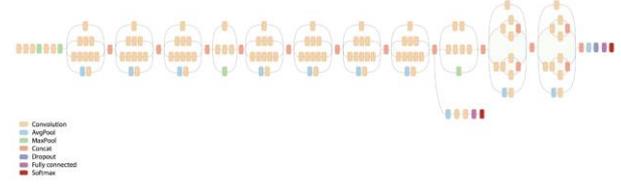

Figure 4. Structure of the convolutional neural network in Google Tensorflow.

The Inception model is compatible with classifying object-based images. However, because of the multi-scale effect, remote-sensing images might not achieve satisfactory results using this method (Zhong et al., 2016). Therefore, in our study, we use two mainstream land-cover data sets (UC-Merced and WHU-SIRI) to retrain the Inception model with a sub-sampling method (LPCNN) introduced by Zhong et al. (2015). In this step, we set a range of size percentages (0.5~1.0) to randomly create subsets of the input images of a certain size. With this method, on the one hand, we can offset the multi-scale effect of remote-sensing imagery. On the other hand, the CNN model can improve its reliability and stability when training and predicting (Krizhevsky et al., 2012; Zhong et al., 2016).

### 3.2. Classifying urban land use at the irregular land-parcel scale

The basic units considered using the proposed method are irregular urban land parcels, which cause difficulties in sampling, feature extraction and training. Therefore, in the proposed method, we sample the sub-land parcels within each parcel on a large scale and input the sample data into the retrained model to obtain the class of each sub-land parcel, which can be regarded as words. The labels of all the sub-land parcels constitute the visual bag-of-words (VBoW), from which word frequencies can be calculated. By inputting the word frequencies into semantic models, the semantic features are generated, which are used to construct an urban land-use classification model (Liu et al., 2017). The procedure for randomly patching the sub-land parcels is as follows:

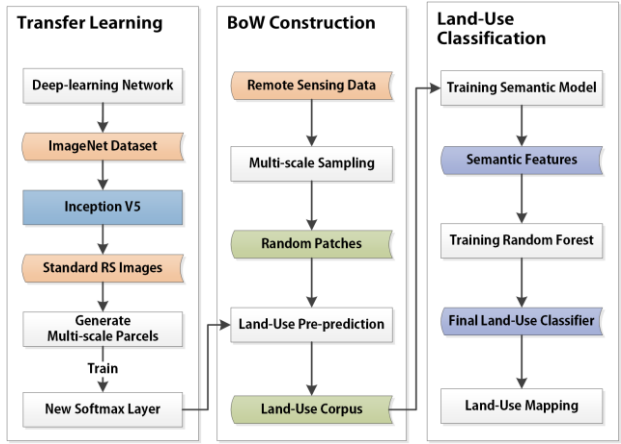

Figure 3. Flowchart of the proposed model for classifying urban land use by integrating Google Tensorflow and scene classification.

### 3.1. Retraining the Google Tensorflow model

Google Inception v5 is a state-of-the-art convolutional neural network (CNN)-based deep-learning model trained on ImageNet object-image-based data sets (Abadi et al., 2016; Krizhevsky et al., 2012) at the ImageNet Challenge, 2015. The structure of Inception v5 is shown in Figure 4. The top-1 and top-5 error rates of this model are 21.2% and 5.6%, respectively, which indicates a perfect image-feature extraction ability (Abadi et al., 2016).

As shown in Figure 4, Inception v5 consists of a convolution layer, average pooling layer, maximum pooling layer, concatenation layer, dropout layer, fully connected layer and Softmax layer. Precisely like the other CNN model, the final Softmax layer can be considered a multi-class logistical classifier whose parameters change through iterations (Abadi et al., 2016; Simonyan and Zisserman, 2014). The former input layer of the Softmax layer is the Bottleneck layer, which can generate the CNN-based features of input images (Jean et al., 2016). After generalization from the pre-trained Inception model, the output high-dimensional vector (with a number of 2048) from the Bottleneck layer can be the best representative vector of any image. Then, the Softmax layer activates the high-dimensional vector and generates the class of each image. The formation of the Softmax function is as follows:

    a. First, we use a remote-sensing image processing module to extract four peripheral coordinates values: $x_{min}, x_{max}, y_{min}, y_{max}$. The combinations of these values can represent the map coordinates of four external rectangle points.
    b. Then, we randomly select the seed point coordinate values $x_{seed}, y_{seed}$ in the range $[x_{min}, x_{max}], [y_{min}, y_{max}]$, respectively, and calculate the smaller value of $x_{max} - x_{seed}$ and $y_{max} - y_{seed}$, which is $l = min(x_{max} - x_{seed}, y_{max} - y_{seed})$.
    c. A constant $w_{min}$ is set as the minimum scale constant of sampling at this time, which means that the

width of each random square window $w \in [w_{min}, l]$. If $l < w_{min}$, then $w = w_{min}$.

d. Considering the continuity and heterogeneity of the land parcels, when the edge of the parcel is over the sample window, an additional judgment is required to determine if the sample window is located in a certain land parcel. In our proposed method, we determine this information by calculating the number of pixels in the sample window that belong to the certain land parcel. If the pixels that belong to a certain land parcel amount to more than 80% of the entire sample window pixels, we regard the sample as valid.

To decrease the uncertainty of random sub-parcel sampling of irregular land parcels, the proposed method repeats the four previously described steps 300 times for each land parcel. Therefore, approximately 1 to 300 different sub-land parcels can be generated in each land parcel. The form of sub-land parcel data in each TAZ is $S = [S_1, S_2, S_3, \cdots]$, where $S_i$ indicates a square parcel whose width belongs to $[w_{min}, l]$ (the unit is 1 pixel). Using the retrained object image-based Inception (v5)-CNN model, we can calculate the word frequencies of the land-use types in each land parcel.

Semantic models are good at mining the topical features within word frequencies. Therefore, they are widely used in land-use scene-classification methods (Liu et al., 2017; Wen et al., 2016; Zhang et al., 2015; Zhong et al., 2015). Because the dictionary size in the proposed method is not large, we adopt a simple TF-IDF transformation, which is a feature-weighted text-topic mining method that is widely used in information retrieval and text mining (Ramos, 2003). The TF-IDF semantic-features transformation method is as follows:

$$\begin{cases} tf_{i,j} = \frac{n_{i,j}}{\sum_k n_{k,j}} \\ idf_i = \log \frac{|D|}{|\{j:t_i \in d_j\}|+1} \end{cases} \quad (2)$$

The $n_{i,j}$ in Equation (2) is the appearance frequency of the $i_{th}$ land-use type in land parcel $d_j$, and $\sum_k n_{k,j}$ represents the total sampling times of a single parcel. $|D|$ represents the number of TAZs in the study area. The denominator of the $log$ function is the number of land parcels that have land use $t_i$. Therefore, the calculation of the TF-IDF value of the $j_{th}$ land-use type in the $i_{th}$ TAZ is as follows:

$$tfidf_{i,j} = tf_{i,j} \times idf_i \quad (3)$$

After calculating the TF-IDF semantic features of each irregular land parcel, we use a random-forest (RFA) and manual-interpretation model to classify the urban land-use type in the study area. RFA is an aggregation of decision-tree classifiers. By extracting random samples from training data sets using the bagging method (Biau, 2012), a new sub-data set is generated. Then, individual decision trees are constructed in each training sub-data set during random feature selection. Unlike the traditional decision-tree method, these decision trees are not pruned during the growth process. Therefore, we can obtain an out-of-bag (OOB)-based estimation error report from each decision tree. By averaging the errors of the decision trees via OOB estimation, the RFA generalization error can be calculated. In certain studies, the RFA-based classification model has overcome the multiple correlative problems among spatial variables, particularly in higher-dimensional fitting situations (Palczewska et al., 2014). Using semantic features and land-use types to train the RFA classifier and test it on the remaining TAZ semantic features from the study area, we obtain the urban land-use types in the study area.

### 3.3. Comparison of methods and accuracy assessment

In this study, accuracy assessment is based on the object-oriented classification evaluation method. By obtaining the final urban land-use classification result and calculating the confusion matrix, we can obtain the overall accuracy (OA) and kappa coefficient, with detailed commission errors, omission errors, product accuracy (PA) and user accuracy (UA) for each class. Additionally, for comparison with the proposed method, we use two different sub-sampling methods to classify the urban land-use type based on the retrained Inception model and the RFA algorithm. These two methods are as follows:

*RECT:* Samples the remote-sensing images in a land parcel's external rectangle shape and directly retrains on the Inception (v5)-CNN model. After the retraining step has been completed, this model is used to classify the urban land-use types at the TAZ level.

*RAND:* Based on the multi-scale sampling method described in Section 3.1, in which the sampling results are directly retrained on the Inception (v5)-CNN model. When predicting the urban land-use types, a voting strategy is used to determine the predominate urban land-use type of a TAZ, whereby the most frequently appearing land-use type is assigned as the TAZ's final land-use type.

## 4. RESULTS

### 4.1. Results of Google Tensorflow retraining

During the retraining the Google Inception model, we created 10,000 and 20,000 remote-sensing parcels with land-cover labels following the completion of multi-scale sampling processes from the UC-Merced and WHU-SIRI land-cover data sets. With these remote-sensing parcels, we constructed the training data set and set the percentages of the training data, validation data and testing data as 0.8, 0.1 and 0.1, respectively. The learning rate of the retraining is set at 0.001 and the number of iterations at 10,000. The batch size of each training step is 100, which means that each time we input 100 samples to fit the model. For the other parameters, we follow the default values.

The code of the retraining model is under the Google Tensorflow framework with GPU boosting. The training and validation accuracy of the retraining process is shown in Figure 5. We can observe that the overall accuracy of retraining the Inception model to classify the urban land-cover type is relatively high (approximately 0.8~0.9). The accuracy converges after 1000 iterations. Compared to WHU-SIRI, retraining on the UC-Merced data set exhibits a higher testing accuracy, which can be explained by the lesser complexity and variety of ground components in the U.S. than in China. (The remote-sensing imagery of UC-Merced is from U.S. locations. WHU-SIRI's locations are in China.) That is, the stronger spatial heterogeneity of the WHU-SIRI data results in lower classification accuracy.

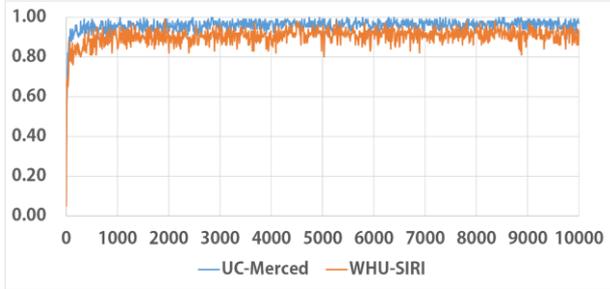

Figure 5. Testing accuracy (y-axis) vs. iteration times (x-axis) during the Google Tensorflow training process.

**4.2. Urban land-use classification results**

When retraining the Google Inception model and fusing the UC-Merced and WHU-SIRI data sets, we set the percentages of training data, validation data and testing data to 0.8, 0.1 and 0.1, respectively. After finishing the retraining process, we obtained an average training accuracy of 88.50% for the testing data set of land-cover data. Based on the method described in Section 3.2, we use a random size square sampling window (minimum size: 20*20 pixels) to calculate the land-cover type frequencies in each TAZ and obtain the TF-IDF-based semantic features for each irregular land parcel. Then, we randomly divide the semantic features data set into two parts: 60% for the training data for RFA-based classifiers and the remaining 40% for testing data. In addition, in the training data set, we select 20% of the data for use in the OOB cross-validation data set.

We applied different methods of sampling and land-use classification, including RECT, RAND and the proposed method. After the training and classifying process was repeated 100 times, the average accuracy-assessment results were as follows. Generally, the classifying accuracy of the proposed method (OA=0.794, Kappa=0.737) was significantly higher than that of traditional sampling and urban land-use classification methods. Only using the external rectangular area of the TAZ to train and classify and ignoring the irregularity of TAZ land parcels, as in the RECT method, brings in external data that can be considered as noise and results in significantly lower classification accuracy (OA=0.140, Kappa=0.060). The RAND method takes the irregularities of land parcels into account and attains higher accuracy than the RECT method (OA=0.528, Kappa=0.418). However, by only considering the voting results, ground objects with small size and a large distribution density (such as Green land) cause errors in classification results. In addition, the RAND method does not consider the high-level semantic information that exists in land parcels, and urban land-use patterns require semantic information involved in scene-classification tasks (Zhong et al., 2015). That is, the RAND method remains unable to bridge the semantic gap, which decreases the quality of the classification results.

| Sampling method | OA | Kappa |
|---|---|---|
| **RECT** | 0.140 | 0.060 |
| **RAND** | 0.528 | 0.418 |
| **Proposed** | 0.794 | 0.737 |

Table 1. Average accuracy of urban land-use classification via different methods.

Figures 6 and 7 and Table 2 display the urban land-use patterns, confusion matrixes and PA/UA of the three sub-land parcel sampling methods, respectively. Based on the results, land parcels with regular shapes, such as Green land, result in the best classification accuracy. The classification accuracy of land parcels with regular shape and low spatial heterogeneity, such as Residential land and Urban village, increases significantly with the RAND method. Last, the RAND method cannot effectively use the semantic information from the remote-sensing images, which results in classification errors regarding Park land and Green land.

From the confusion matrix (Figure 7), we can observe that the proposed method can effectively increase classification accuracy and minimize the confusion condition between different ground components. For example, the classification accuracy for Residential land is the highest: 90.16%. Because Residential land is typically adjacent to Commercial land and Public-management-services land, high-level semantic features must be considered in classfying these land-use types. Therefore, in the RAND and RECT sub-sampling methods, the classification accuracies for Residential land are low. By integrating the deep-learning method, the proposed method could effectively extract high-level semantic features and obtain highly accurate land-use classification results.

Notably, while Green land and Park land could be effectively distinguished, 11.76% of Park land was mistakenly classified as Green land. Urban village is a ground component that consists of patches crowded with houses. Its classification accuracy reached 71.31%, with 18.03% mistakenly classified as Residential land. To our knowledge, Green land has little visual difference from Park land. Similarly, Urban village differs little in appearance from Residential land. That is, only by probing deeply into the inner structure of population characteristics could the method discern the differences between visually similar remote-sensing image parcels. From remote-sensing imagery, we can only obtain natural physical properties, which is inadequate for urban land-use detection because the socio-economic properties are lacking (Hu et al., 2016; Liu et al., 2017), thus resulting in classification errors.

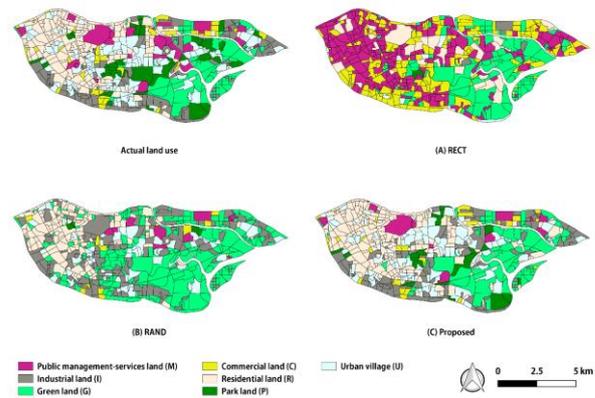

Figure 6. Urban land-use classification results for the different methods: (A) RECT, (B) RAND, (C) Proposed method.

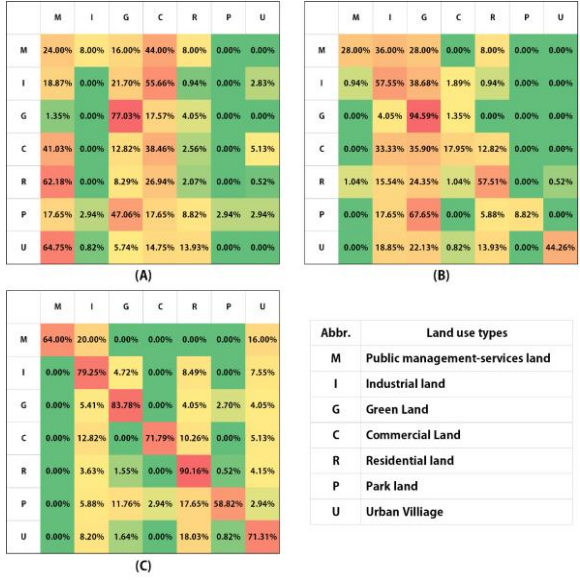

Figure 7. Confusion matrixes of urban land-use classification results for the different methods: (A) RECT; (B) RAND and (C) Proposed method

| Sampling method | LU | Commission error | Omission error | Product accuracy | User accuracy |
|---|---|---|---|---|---|
| **RECT** | M | 0.760 | 0.976 | 0.024 | 0.240 |
| | I | 1.000 | 1.000 | 0.000 | 0.000 |
| | G | 0.230 | 0.555 | 0.445 | 0.770 |
| | C | 0.615 | 0.914 | 0.086 | 0.385 |
| | R | 0.979 | 0.871 | 0.129 | 0.021 |
| | P | 0.971 | 0.000 | 1.000 | 0.029 |
| | U | 1.000 | 1.000 | 0.000 | 0.000 |
| **RAND** | M | 0.720 | 0.300 | 0.700 | 0.280 |
| | I | 0.425 | 0.579 | 0.421 | 0.575 |
| | G | 0.054 | 0.694 | 0.306 | 0.946 |
| | C | 0.821 | 0.462 | 0.538 | 0.179 |
| | R | 0.425 | 0.196 | 0.804 | 0.575 |
| | P | 0.912 | 0.000 | 1.000 | 0.088 |
| | U | 0.557 | 0.018 | 0.982 | 0.443 |
| **Proposed Method** | M | 0.360 | 0.000 | 1.000 | 0.640 |
| | I | 0.208 | 0.282 | 0.718 | 0.792 |
| | G | 0.162 | 0.184 | 0.816 | 0.838 |
| | C | 0.282 | 0.034 | 0.966 | 0.718 |
| | R | 0.098 | 0.202 | 0.798 | 0.902 |
| | P | 0.412 | 0.167 | 0.833 | 0.588 |
| | U | 0.287 | 0.230 | 0.770 | 0.713 |

Table 2. Urban classification accuracy of each land-use type for the different methods.

### 4.3. Parameter sensitivity analysis

Using the retrained Inception model (based on the fused UC-Merced and WHU-SIRI data) and constructing effective TF-IDF semantic features, we attained an overall accuracy of 0.794 for the proposed study area. The random-minimum scale parameter $w_{min}$ is sensitive when assigning different values. This parameter is used because we require a sampling window with a certain size to extract the sub-land parcels and use the retrained model to assign a label to this sub-land parcel. Therefore, we tested the sample-window size with different values using the two land-cover data sets (UC-Merced and WHU-SIRI). The size of the sub-land parcels ranged from 10 to 100 pixels. The final accuracy results are presented in Figure 8. We could calculate that the overall accuracy first increases and then decrease with the increase in $w_{min}$ for the two land-cover data sets. The edge effect (i.e., that the edged parcel covers a substantial number of outlier data and a large amount of noise) causes erroneous words to appear in the VBoW, thus decreasing the accuracy of the urban land-use classification results. Therefore, the proposed method adopts a $w_{min}$ of 20 pixels to obtain the best classification results.

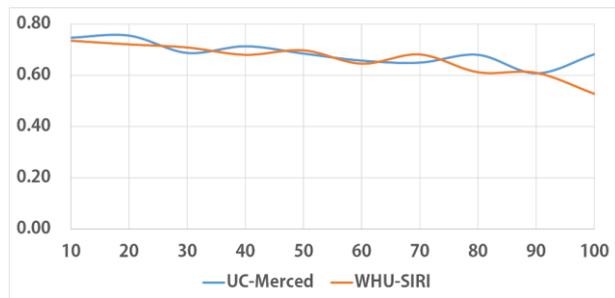

Figure 8. Accuracy assessment (y-axis) of urban land-use classification using different sampling windows (x-axis) during prediction processes.

## 5. DISCUSSION AND CONCLUSIONS

The question of how to improve urban land-use classification results for the moderate scale has been a popular research topic in the fields of remote sensing and GIS (Chen et al., 2017; Liu et al., 2017; Yao et al., 2016; Yuan et al., 2012). This study proposed an effective model to classify urban land use at the irregular land-parcel level by integrating a convolutional neural network trained on object-based images (Inception). First, we used public land-cover data sets to retrain the high-accuracy Inception (v5)-CNN model on public land-cover datasets. We used this retrained model to generate the classes of sub-land parcels within each irregular land parcel (TAZ). Then, we calculated the class frequencies of these sub-land parcels in each TAZ and constructed a VBoW by adopting TF-IDF, a simple semantic model used to generate high-dimension semantic features. Subsequently, we trained an RFA classifier to classify the urban land-use patterns in the study area (OA=0.793, Kappa=0.737).

Compared to traditional urban land-use classification methods, the proposed model, which integrates deep learning and semantic classification, could accurately classify the urban land-use types at the level of irregular land parcel. The retraining data for Inception (v5)-CNN were public land-cover data sets (UC-Merced and WHU-SIRI) downloaded from the Internet, which indicates the transfer-learning capability of the proposed model to a certain extent. By integrating a semantic model, we could mine the semantic information from high-resolution remote-sensing image data at the irregular land-parcel level and classify the urban land-use type based on scene classification. In future research, more effective, classical probabilistic topic models (PTMs), such as pLSA, LDA and Word2Vec (Yao et al., 2016), should be considered to effectively mine high-level semantic information and increase our model's accuracy.

This study introduced a method that integrates deep learning and a semantic model to mine the urban functional structure. However, this method has several inadequacies. The data used in the proposed method were only high-resolution remote-sensing images, which can only be used to mine the natural physical information of ground components. The inability to mine socio-economic information inside the land parcels causes a classification error, for example, regarding ground objects with similar visual effects, such as Green land and Park land. Liu et al. (2017) proposed a method that fuses HSR images and social media data and attained relatively high accuracy (OA=0.865, Kappa=0.828) in the same study area. However, they attained lower accuracy when only remote-sensing images were used (OA=0.685, Kappa=0.591). Therefore, in future research, we should consider adding multi-source social media data to the model and seek to effectively mine the socio-economic properties while increasing urban land-use classification accuracy. Additionally, automatically generating sub-land parcels in irregular land parcels is a topic worthy of further study. The sampling method in the proposed model is relatively simple and could not obtain complete sub-land parcels in certain long, broken TAZs, which resulted in classification errors. This problem will be discussed in future research.

Generally, the proposed model could effectively overcome the multi-scale effect of high-resolution remote-sensing images, classify urban land-use types at the level of irregular land parcels and demonstrate the ability of transfer learning. In future research, we will apply multi-source social media data to fully consider socio-economic properties. Thus, we hope to obtain more accurate urban functional structure and urban land-use classification results. The proposed model can meet the future need for the rapid detection of urban functional zones and urban land-use types, thus effectively supporting urban planning and government management.

### ACKNOWLEDGEMENTS

This study was supported by National Key R&D Program of China (Grant No.2017YFA0604402), the Key National Natural Science Foundation of China (Grant No. 41531176) and the National Natural Science Foundation of China (Grant No. 41671398).